\documentclass[conference]{IEEEtran}
\IEEEoverridecommandlockouts

\usepackage[T1]{fontenc}
\usepackage[utf8]{inputenc}
\usepackage{amsmath,amssymb,amsfonts}
\usepackage{graphicx}
\usepackage{booktabs}
\usepackage{multirow}
\usepackage{array}
\usepackage{xcolor}
\usepackage{textcomp}
\usepackage{pifont}
\usepackage{url}
\usepackage[hidelinks,breaklinks]{hyperref}
\usepackage{xurl}   

\newcommand{\fv}[1]{\texttt{\detokenize{#1}}}

\newcommand{\ie}{i.e.,}
\definecolor{hl}{HTML}{B3202C}

\makeatletter
\def\abstract{\par\noindent\small\textbf{\textit{Abstract:}}\ \ignorespaces}

\makeatother

\begin{document}

\title{Accurate Ensembles, Fragile Narratives:\\
Multi-Scale Stacking and a Fidelity Audit of\\
LLM-Generated Explanations for Credit Risk}

\author{%
\IEEEauthorblockN{Gregorius Reynaldi Pratama}
\IEEEauthorblockA{\textit{School of Computer Science and Technology}\\
\textit{Harbin Institute of Technology, Shenzhen}\\
Shenzhen, China \\
gregoriusreynaldi@gmail.com}
\and
\IEEEauthorblockN{Kuo-Kun Tseng}
\IEEEauthorblockA{\textit{School of Computer Science and Technology}\\
\textit{Harbin Institute of Technology, Shenzhen}\\
Shenzhen, China \\
kktseng@hit.edu.cn}
}

\maketitle

\begin{abstract}
Credit scoring increasingly relies on model families whose decision logic
cannot be read off their parameters, which places them in tension with
supervisory expectations that adverse decisions be explainable. A common
proposal is to close that gap with a language model: compute feature
attributions, hand them to an LLM, and let it write the rationale. We build
such a system end to end and then test whether the second half of the promise
holds. The predictive component is a multi-scale stacking ensemble that fuses
four differently regularised gradient-boosting learners with a residual
network through a small neural meta-learner trained on out-of-fold
predictions. On a public 32{,}581-application credit dataset it reaches a test
ROC-AUC of $0.9539$ (95\% CI $[0.9462,0.9616]$) and PR-AUC of $0.9137$,
improving on the strongest single model by $\Delta\mathrm{AUC}=0.0143$; the
improvement remains significant ($p=0.016$) even under a deliberately
conservative independence assumption. The explanation component fuses SHAP and
LIME attributions and passes only the top-ranked drivers to a
2.7B-parameter model under greedy decoding. Our central finding is
asymmetric. The ranking gains are real but operationally small: at the
$F_1$-optimal threshold the ensemble avoids only six additional missed
defaults out of $1{,}422$ relative to a tuned random forest, cutting
cost-weighted loss by under $2\%$. The narrative layer, in contrast, fails in
a way that prompt engineering alone does not fix. In an audited case the model
named three factors as risk-increasing that the underlying attributions
scored as risk-\emph{reducing}. We trace this to properties we measure rather
than assume: SHAP and LIME agree on \emph{which} features matter
(overlap@10 $=0.80$) but not on their order ($\tau=0.43$, $p=0.18$), and the
sign of the attribution for a feature such as applicant income is close to a
coin flip across instances (modal-sign share $0.53$). We also report
calibration ($\mathrm{ECS}=0.117$) and perturbation stability
($\mathrm{DPD}=0.078$) that both fall short of the thresholds our own protocol
declares acceptable. We conclude that constrained prompting is a necessary but
not sufficient control, and that grounding must be verified after generation
rather than assumed from the prompt.
\end{abstract}


\section{Introduction}

A credit decision is a decision about a person. When a lender declines an
application or prices it punitively, the applicant in most jurisdictions
acquires a right to be told why, and the supervisor acquires an interest in
whether the stated reason is the operative one~\cite{goodman2017eu}. This is
what distinguishes credit scoring from most tabular classification problems:
the score is not the deliverable. The score plus a defensible account of the
score is the deliverable.

The modelling literature has largely optimised the first half. Gradient-boosted
trees and their ensembles now dominate credit benchmarks~\cite{lessmann2015,
dastile2020}, and the margin over interpretable baselines such as regularised
logistic regression is large enough that institutions are unwilling to give it
up. The standard response is post-hoc attribution: fit whatever model performs
best, then explain it with SHAP~\cite{lundberg2017} or LIME~\cite{ribeiro2016}.
This has become close to a default in credit risk management~\cite{bussmann2021}.

Attribution, however, produces a vector of signed real numbers, one per
feature, expressed in units of log-odds or probability shift. A loan officer
does not read that vector, and neither does an applicant. A recent and fast
growing line of work therefore proposes a further translation step: feed the
attributions to a large language model and let it write prose~\cite{
zytek2024llmxai, mavrepis2024xaiforall, kroeger2023posthoc, stories2026,
creditllm2026}. The appeal is obvious. The risk is equally obvious, because
language models generate fluent text whether or not it is grounded in what
they were given~\cite{ji2023hallucination}, and fluency is precisely what makes
an ungrounded credit rationale dangerous: it is more persuasive than the
correct one, not less.

Most existing proposals control this risk at the input. They restrict the
prompt to a handful of top-ranked drivers, forbid the model from inventing
numbers, and decode greedily. We implement exactly this design and then ask
the question that the design's advocates usually leave to future work:
\emph{does it work?} Not whether the output reads well, since it plainly
does, but whether the propositions it asserts match the attributions it was
handed.

The answer, on our system, is no. This paper reports that result together with
the diagnostic work needed to explain it, and it revises the accompanying
predictive claims downward where the evidence requires.

\subsection*{Contributions}

\begin{enumerate}
\item \textbf{A multi-scale stacking ensemble for credit default prediction.}
Four gradient-boosting learners deliberately configured at different
depth/learning-rate scales, plus a residual network, are fused by a small
neural meta-learner trained on out-of-fold predictions. It attains test
ROC-AUC $0.9539$ and PR-AUC $0.9137$ (Section~\ref{sec:results}).

\item \textbf{Statistical rather than rhetorical model comparison.} We attach
confidence intervals to every headline metric and test the ensemble's margin
over the best single model under an assumption chosen to disadvantage our own
claim. The margin survives ($p=0.016$).

\item \textbf{An operating-point analysis that qualifies the headline.} We show
that the ranking gain translates into a cost reduction of under $2\%$ at the
deployed threshold across a wide range of misclassification cost ratios
(Section~\ref{sec:cost}). Reporting AUC alone would have overstated the
practical benefit.

\item \textbf{A measured account of what the attribution layer actually
provides.} The ensemble's risk representation differs qualitatively from that
of the linear baseline: \fv{loan_grade} absorbs the role that
\fv{loan_int_rate} and the bureau default flag play in logistic regression
(Section~\ref{sec:explanations}). We quantify SHAP--LIME agreement,
cross-model agreement, attribution concentration and directional stability
instead of asserting them.

\item \textbf{A fidelity audit of the narrative layer, with a documented
failure.} We define a lightweight sign-and-membership audit, apply it, and
report a case in which the generated rationale inverts the direction of three
of the four drivers it was given (Section~\ref{sec:fidelity}). We connect this
to the attribution instability measured in the previous section rather than
treating it as a one-off.

\item \textbf{Honest trustworthiness diagnostics.} Calibration, perturbation
stability, local explanation consistency and group disparity are reported
against the acceptance thresholds our own protocol specifies. Three of the
four fail (Section~\ref{sec:trust}). We argue this is more useful to a
practitioner than a table of green ticks.
\end{enumerate}

A note on provenance is in order. An earlier version of this work reported a
test ROC-AUC of $0.96$. Recomputation from the persisted artefacts gives
$0.9539$. Every figure reported in this paper is recomputed from the serialised
experiment outputs, and the reproduction scripts are released with the code.

\section{Related Work}

\subsection{Machine learning for credit scoring}

Benchmarking studies over the past decade converge on a consistent ordering:
ensembles of decision trees outperform single learners and outperform neural
networks on typical credit-scoring feature sets, with the margin over
regularised logistic regression material but not
overwhelming~\cite{lessmann2015, dastile2020}. The broader tabular literature
reports the same phenomenon and offers explanations grounded in the geometry of
tabular decision boundaries: tree ensembles are biased towards axis-aligned,
piecewise-constant functions, which suits features such as grade bands and
debt-ratio thresholds~\cite{grinsztajn2022, shwartzziv2022}. Our results
reproduce this ordering, with random forest and XGBoost outperforming every
neural architecture we test, and our ensemble is designed around it rather
than against it, using the neural component as one voice among five rather
than as a replacement.

\subsection{Post-hoc attribution and its limits}

SHAP~\cite{lundberg2017} grounds attribution in the Shapley
value~\cite{shapley1953, strumbelj2014}, which uniquely satisfies local
accuracy, consistency and missingness. LIME~\cite{ribeiro2016} fits a sparse
linear surrogate in a sampled neighbourhood. Both are model-agnostic and both
are widely deployed in finance~\cite{bussmann2021, guidotti2018, arrieta2020}.

Their limitations are also well documented. LIME's reliance on random
perturbation makes its explanations unstable across
runs~\cite{alvarezmelis2018}, and both methods can be deliberately fooled by an
adversary who exploits the off-manifold queries they
issue~\cite{slack2020fooling}. Rudin argues from these failure modes that
high-stakes decisions should use models that are interpretable by
construction~\cite{rudin2019}. We do not resolve that debate. We do take from
it the methodological point that an attribution vector is evidence about a
model, not a transcript of it, and that anything built downstream inherits its
noise. Section~\ref{sec:explanations} measures that noise on our system;
Section~\ref{sec:fidelity} shows it propagating.

\subsection{Language models as explanation interfaces}

The proposal to have an LLM verbalise XAI output is recent and
active~\cite{zytek2024llmxai, mavrepis2024xaiforall}. Kroeger et
al.~\cite{kroeger2023posthoc} ask whether LLMs can act as post-hoc explainers
directly. Factorial studies of narrative quality find that readers prefer
LLM-written explanations to raw attribution
tables~\cite{stories2026}, which is the intended effect and also the reason a
wrong narrative is worse than a wrong table. Work specific to credit
risk asks whether LLMs can serve as governance-grade explainability
tools~\cite{creditllm2026} and whether LLM-derived feature rationales align
with those of classical models~\cite{creditalign2025}. Two-stage designs that
add an explicit verification pass over the generated
text~\cite{twostage2026} are, in our reading, the correct architectural
response, and our results are an empirical argument for them.

Faithfulness, meaning whether an explanation reflects the reasoning it purports
to describe, has a precise treatment in NLP~\cite{jacovi2020faithfulness},
and hallucination in conditional generation is
surveyed in~\cite{ji2023hallucination}. Our contribution here is narrow and
concrete: we apply a minimal faithfulness check to a credit-risk narrative
pipeline built the way the literature recommends, and report that it fails.

\subsection{Positioning}

Relative to prior work, the novelty of this paper is not the ensemble, and it
is not the idea of pairing attribution with an LLM. It is the audit. We are
not aware of a credit-risk study that reports both a full attribution-stability
profile of its own ensemble and a documented sign-inversion failure of its own
narrative layer, and that connects the two.

\section{Data and Preprocessing}
\label{sec:data}

\subsection{Dataset}

We use the public Credit Risk Dataset, comprising $32{,}581$ consumer loan
applications described by eleven predictors and a binary outcome
\fv{loan_status} ($1=$ default). Seven predictors are numeric and four
categorical; Table~\ref{tab:features} lists them. The target is imbalanced:
$25{,}473$ non-defaults ($78.18\%$) against $7{,}108$ defaults ($21.82\%$), a
ratio of $3.58{:}1$.

Overall completeness is $98.97\%$. Missingness is confined to two continuous
variables, \fv{loan_int_rate} ($3{,}116$ records, $9.56\%$) and
\fv{person_emp_length} ($895$ records, $2.75\%$), and is scattered rather than
structured. Duplicate rows account for $165$ records ($0.51\%$).

\begin{table}[t]
\caption{Predictors and target. Names are given exactly as they appear in the
released data and code.}
\label{tab:features}
\centering
\scriptsize
\setlength{\tabcolsep}{4pt}
\begin{tabular}{@{}>{\ttfamily}l p{0.40\columnwidth}@{}}
\toprule
\normalfont\textbf{Feature} & \textbf{Description}\\
\midrule
\multicolumn{2}{@{}l}{\normalfont\itshape Numeric predictors}\\
person\_age                  & Applicant age in years\\
person\_income               & Annual income\\
person\_emp\_length          & Employment tenure in years\\
loan\_amnt                   & Requested principal\\
loan\_int\_rate              & Contractual interest rate\\
loan\_percent\_income        & Principal as a share of income\\
cb\_person\_cred\_hist\_length & Length of credit history\\
\addlinespace[2pt]
\multicolumn{2}{@{}l}{\normalfont\itshape Categorical predictors}\\
person\_home\_ownership      & Rent, mortgage, own or other\\
loan\_intent                 & Stated purpose, six levels\\
loan\_grade                  & Internal grade A to G, ordinal, A best\\
cb\_person\_default\_on\_file & Prior bureau default, binary\\
\addlinespace[2pt]
\multicolumn{2}{@{}l}{\normalfont\itshape Target}\\
loan\_status                 & \normalfont$1=$ default, $0=$ repaid\\
\bottomrule
\end{tabular}
\end{table}

\subsection{Leakage control}

The single most consequential preprocessing decision is ordering. We partition
before we learn anything from the data. The design matrix and labels are split
$80/20$ with \texttt{stratify=y}, preserving the $21.82\%$ default rate in both
partitions, yielding $26{,}064$ training and $6{,}517$ test records
($5{,}095$ non-defaults, $1{,}422$ defaults). Every subsequent transformation
(winsorisation caps, imputation, scaling, encoding) estimates its parameters
on the training partition only and is then applied unchanged to the
test partition. The fitted transformer objects are serialised so that the
inference service applies numerically identical transformations.

\subsection{Transformations}

\textbf{Winsorisation.} Domain-implausible extremes were capped rather than
deleted: \fv{person_age} at $100$ years and \fv{person_emp_length} at $50$
years, against raw maxima of $144$ and $123$ respectively. Caps are estimated
on training data and reused.

\textbf{Imputation.} The two incomplete numeric columns are filled with a
$k$-nearest-neighbour imputer ($k=5$). Because the missing variables are
correlated with income and principal, local interpolation is preferable to
global mean or median substitution.

\textbf{Scaling.} Numeric features are centred on the training median and
scaled by the training interquartile range. Income is severely right-skewed
(sample skewness $\approx 32$, driven by a small number of very high earners),
which is exactly the regime in which a robust scaler outperforms
standardisation for gradient-based learners.

\textbf{Encoding.} \fv{loan_grade} carries a natural order (A $<\cdots<$ G) and
is ordinal-encoded so that higher values mean worse credit quality.
\fv{person_home_ownership} and \fv{loan_intent} are one-hot encoded with the
first level dropped and \texttt{handle\_unknown='ignore'} so that unseen
categories at inference yield a zero vector rather than an exception.
\fv{cb_person_default_on_file} is mapped $\{$N,Y$\}\!\to\!\{0,1\}$.

Concatenation yields a $17$-dimensional design matrix used identically by every
model reported here.

\section{Multi-Scale Stacking Ensemble}
\label{sec:method-model}

\begin{figure*}[t]
\centering
\includegraphics[width=\textwidth]{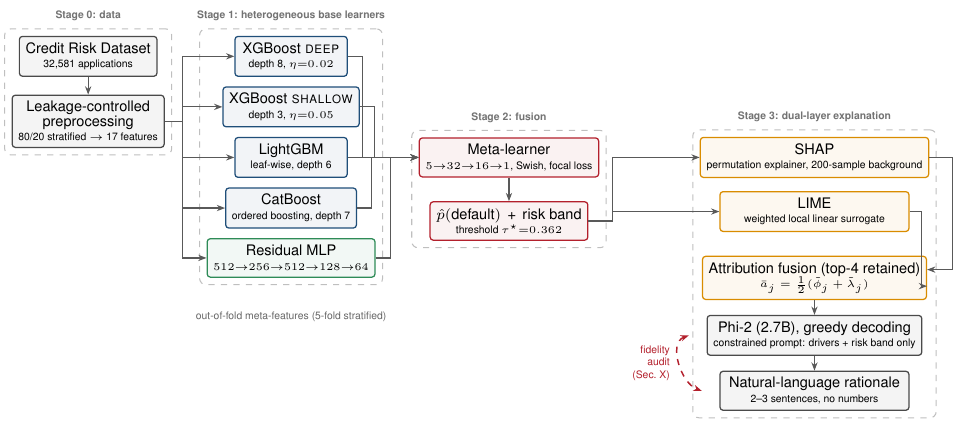}
\caption{End-to-end system. Five heterogeneous base learners are fused by a
neural meta-learner trained on out-of-fold predictions; the resulting
probability drives both the credit decision and the explanation stack. The
dashed edge marks the interface audited in Section~\ref{sec:fidelity}, where
the attributions handed to the language model are compared against the
propositions it asserts.}
\label{fig:arch}
\end{figure*}

\subsection{Design rationale}

Stacked generalisation~\cite{wolpert1992} improves on a single learner when the
base learners make \emph{different} mistakes. Diversity is therefore the design
objective, not accuracy of the individual components. Credit data is mixed-type,
moderately imbalanced, and contains both broad monotone trends (higher
repayment burden, higher risk) and sharp local interactions (a marginal grade
combined with short tenure). We assemble learners whose inductive biases cover
that range.

\textbf{Multi-scale gradient boosting.} Rather than four tuned variants of the
same configuration, we deliberately spread the learners across the
depth--shrinkage plane. A deep XGBoost (depth $8$, $\eta=0.02$, $500$ trees)
captures high-order interactions. A shallow XGBoost (depth $3$, $\eta=0.05$,
$300$ trees) acts as a low-variance anchor that represents global monotone
structure. LightGBM's leaf-wise growth~\cite{ke2017lightgbm} concentrates
capacity on high-gradient regions, improving coverage of sparse segments.
CatBoost's ordered boosting~\cite{prokhorenkova2018catboost} reduces target
leakage in categorical splits, which matters for the one-hot grade and intent
blocks. Hyperparameters are given in Table~\ref{tab:base}.

\textbf{Residual network.} Tree learners approximate smooth functions with
staircases. The fifth base learner is a residual MLP over the scaled numeric
manifold: dense layers of $512$ and $256$ units, a residual block expanding
back to $512$ with an identity skip~\cite{he2016resnet}, then $128$ and $64$
units before a sigmoid output. It contributes a smooth approximator whose
errors are, by construction, poorly correlated with the trees'.

\begin{table}[t]
\caption{Base-learner configurations. The four boosting models are spread
across the depth--shrinkage plane by design rather than tuned to a common
optimum.}
\label{tab:base}
\centering
\footnotesize
\setlength{\tabcolsep}{4pt}
\begin{tabular}{@{}lccccc@{}}
\toprule
\textbf{Learner} & \textbf{Depth} & \textbf{Trees} & $\boldsymbol{\eta}$ &
\textbf{Subsample} & \textbf{Colsample}\\
\midrule
XGBoost \textsc{deep}    & 8 & 500 & 0.020 & 0.80 & 0.80\\
XGBoost \textsc{shallow} & 3 & 300 & 0.050 & 0.90 & 0.90\\
LightGBM                 & 6 & 400 & 0.030 & 0.85 & 0.85\\
CatBoost                 & 7 & 450 & 0.025 & 0.80 & 0.80\\
Residual MLP             & n/a & n/a & 0.001 & n/a & n/a\\
\bottomrule
\end{tabular}
\end{table}

\subsection{Imbalance handling}

Every base learner absorbs the class imbalance before stacking, so that the
meta-learner receives comparably scaled signals. The boosting models use
\begin{equation}
w_{+} \;=\; \frac{1-\pi}{\pi} \;=\; \frac{1-0.2182}{0.2182} \;\approx\; 3.58 ,
\end{equation}
with $\pi$ the training default rate. The neural components use focal
loss~\cite{lin2017focal},
\begin{equation}
\mathcal{L}_{\mathrm{focal}}(p_t) \;=\; -\,\alpha_t\,(1-p_t)^{\gamma}\log p_t ,
\end{equation}
with $\alpha=0.25$ for the positive class, $0.75$ otherwise, and $\gamma=2.0$.
Focal loss down-weights the abundant easy negatives and concentrates gradient
on the hard positives, which is the behaviour we want when defaults are the
minority and the expensive error.

\subsection{Out-of-fold meta-features}

\begin{figure*}[t]
\centering
\includegraphics[width=0.92\textwidth]{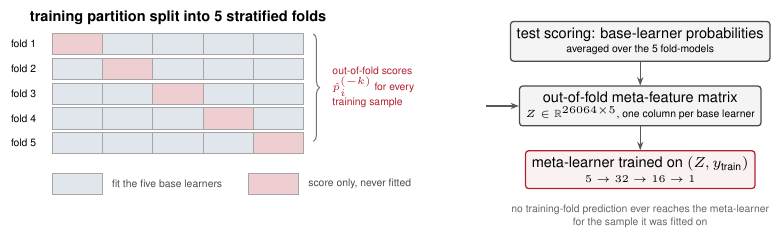}
\caption{Out-of-fold construction of the meta-feature matrix. Within each of
five stratified folds the base learners are refitted on the complementary four
folds and score only the held-out block, so no meta-feature is ever produced by
a model that saw the corresponding label. Test-set meta-features average the
five fold-models.}
\label{fig:oof}
\end{figure*}

Training a meta-learner on in-sample base predictions leaks the target: the
base models are overconfident on data they memorised, and the meta-learner
learns to trust that overconfidence. We avoid this with the standard
out-of-fold construction, illustrated in Fig.~\ref{fig:oof}.

Let $\mathcal{D}_{\mathrm{tr}}$ be partitioned into five stratified folds
$\{\mathcal{F}_k\}_{k=1}^{5}$. For each base learner $m$ and fold $k$ we refit
on $\mathcal{D}_{\mathrm{tr}}\setminus\mathcal{F}_k$ and score $\mathcal{F}_k$:
\begin{equation}
Z_{i,m} \;=\; \hat p_{m}^{(-k(i))}\!\left(\mathbf{x}_i\right),
\qquad i \in \mathcal{F}_{k(i)} ,
\end{equation}
where $k(i)$ is the fold containing sample $i$. The result is a meta-feature
matrix $Z \in \mathbb{R}^{26064 \times 5}$ in which every entry is an honest
held-out probability. Test meta-features average the five fold-models,
\begin{equation}
Z^{\mathrm{te}}_{i,m} \;=\; \frac{1}{5}\sum_{k=1}^{5}
\hat p_{m}^{(-k)}\!\left(\mathbf{x}_i\right) ,
\end{equation}
which also reduces variance relative to any single refit.

\subsection{Meta-learner}

The meta-learner is deliberately small: with only five inputs, capacity is a
liability. It maps $5 \to 32 \to 16 \to 1$ with Swish
activations~\cite{ramachandran2017swish}, $L_2$ penalty $10^{-4}$, batch
normalisation~\cite{ioffe2015batchnorm} after each hidden layer, and dropout
$0.2$~\cite{srivastava2014dropout} after the first. Optimisation uses
Adam~\cite{kingma2015adam} at $10^{-3}$ with focal loss, plateau-triggered
learning-rate halving (factor $0.5$, patience $5$, floor $10^{-6}$) and early
stopping on validation ROC-AUC (patience $10$, best weights restored). The
narrow $32\!\to\!16$ bottleneck forces compact combination rules; batch
normalisation stabilises training on the strongly correlated meta-features.

Monitoring ROC-AUC rather than loss is a deliberate choice under imbalance:
validation loss can improve while ranking quality degrades, and ranking is what
the downstream threshold consumes.

\subsection{Threshold selection}

A $0.5$ cut-off is inappropriate at a $21.8\%$ base rate. We select
\begin{equation}
\tau^{\star} \;=\; \arg\max_{\tau}\;
\frac{2\,\mathrm{Prec}(\tau)\,\mathrm{Rec}(\tau)}
     {\mathrm{Prec}(\tau)+\mathrm{Rec}(\tau)}
\end{equation}
on training data, giving $\tau^{\star}=0.362$ for the ensemble, and apply it
unchanged to the test partition. We stress in Section~\ref{sec:cost} that
$F_1$ is a convenient but economically arbitrary criterion, and that the choice
of $\tau$ matters more to realised cost than the choice of model.

\section{Dual-Layer Explanation Framework}
\label{sec:method-xai}

\subsection{Layer 1: quantitative attribution}

\textbf{SHAP.} Because the ensemble is a composition of five heterogeneous
learners and a network, no model-specific fast path applies. We wrap the entire
pipeline (base scoring, stacking, meta-learner) in a single callable and
explain \emph{that}, using a permutation explainer with a $200$-sample
background drawn from training data. Attributions are computed for $100$ test
instances, giving $\Phi \in \mathbb{R}^{100\times 17}$. Explaining the
composition rather than a component is the methodologically important part:
attributions taken from one base learner would describe a model that is not the
one making the decision. Global importance is $\bar\phi_j = \frac{1}{n}\sum_i
|\phi_{ij}|$.

\textbf{LIME.} For each of $50$ test instances we fit a proximity-weighted
sparse linear surrogate on perturbed neighbours,
\begin{equation}
\begin{aligned}
g^{\ast} &= \arg\min_{g\in\mathcal{G}}\;
   \sum_{z}\pi_{\mathbf{x}}(z)\bigl(f(z)-g(z)\bigr)^{2}+\Omega(g),\\[2pt]
\pi_{\mathbf{x}}(z) &= \exp\!\bigl(-\lVert z-\mathbf{x}\rVert^{2}/\sigma^{2}\bigr),
\end{aligned}
\end{equation}
and take the surrogate coefficients as local importances.

\textbf{Fusion.} SHAP and LIME scores are min--max normalised to a common
scale, $\tilde\phi_j$ and $\tilde\lambda_j$, and averaged,
$\bar a_j = \tfrac{1}{2}(\tilde\phi_j + \tilde\lambda_j)$. The intent is that
consensus suppresses method-specific artefacts. Section~\ref{sec:explanations}
examines whether the two methods agree enough for this averaging to be
meaningful.

\subsection{Layer 2: narrative generation}

The narrative layer uses \texttt{microsoft/phi-2}, a 2.7B-parameter decoder-only
transformer~\cite{phi2}, loaded in \texttt{float32} with left-side padding.
Model choice is driven by deployment constraints: narrative generation is an
inference-time cost per application, and a 2.7B model runs on the same
commodity hardware as the scoring service.

The prompt is deliberately impoverished. It contains the predicted probability,
a discretised risk band (\textsc{high}/\textsc{moderate}/\textsc{low}), the
raw applicant profile, the top four fused drivers with signed impacts, the
five globally most important features, and the three most sensitive features.
It instructs the model to write two to three sentences, to reference at least
three named features, to describe how they change risk, and to avoid both
jargon and numerals. Decoding is greedy (\texttt{do\_sample=False},
\texttt{repetition\_penalty=1.05}). The full template is reproduced in the
appendix.

Every control here is an \emph{input-side} control. Nothing in the design
inspects the generated text. This is the standard configuration in the
literature, and Section~\ref{sec:fidelity} is an argument that it is
insufficient.

\section{Experimental Protocol}
\label{sec:protocol}

\subsection{Metrics}

Accuracy is excluded. At a $21.8\%$ default rate a constant non-default
predictor achieves $78\%$ accuracy while providing no risk discrimination, so
accuracy would reward exactly the behaviour we need to detect. We report
ROC-AUC for threshold-free ranking quality, PR-AUC for behaviour on the
minority class, and $F_1$, precision, recall and specificity at the selected
operating point. Section~\ref{sec:cost} adds a cost-weighted view.

\subsection{Uncertainty quantification}

Point estimates without intervals invite over-reading of small differences. We
report a $95\%$ confidence interval for every AUC using the Hanley--McNeil
variance approximation~\cite{hanley1982},
\begin{equation}
\begin{split}
\widehat{\mathrm{Var}}(A)=\frac{1}{n_{+}n_{-}}\Bigl[\,
  &A(1-A)+(n_{+}-1)(Q_{1}-A^{2})\\[-1pt]
  &+\,(n_{-}-1)(Q_{2}-A^{2})\Bigr],
\end{split}
\end{equation}
with $Q_1 = A/(2-A)$ and $Q_2 = 2A^2/(1+A)$, and Wilson score
intervals~\cite{wilson1927} for recall, precision and specificity, which
behave correctly near the boundary where the normal approximation does not.

For the ensemble-versus-baseline comparison the correct instrument is a
correlated test such as DeLong's~\cite{delong1988}, which requires the paired
score vectors. Those were not retained by the original experiment. We therefore
compute the $z$ statistic under the assumption that the two AUCs are
\emph{independent}. Since predictions on a shared test set are strongly
positively correlated, and
$\mathrm{Var}(A_1-A_2)=\sigma_1^2+\sigma_2^2-2\rho\sigma_1\sigma_2$ decreases
in $\rho$, this inflates the standard error and understates significance. Any
conclusion that survives it would survive the correct test. We report the
sensitivity to $\rho$ explicitly rather than quietly assuming a favourable
value.

\subsection{Reproducibility}

Experiments ran on Windows 11 with an Intel Core Ultra 9 275HX (24 cores),
32\,GB RAM, and GPU acceleration, under Python 3.10.13 with NumPy 1.26, pandas
2.1, SciPy 1.11, scikit-learn 1.4, XGBoost 2.0, LightGBM 4.0, CatBoost 1.2,
TensorFlow 2.15, PyTorch 2.0 with pytorch-tabnet 4.0, SHAP 0.45 and LIME 0.2.
A fixed seed, 5-fold stratified cross-validation and the $80/20$ stratified
split are enforced throughout. Training and test tensors, fitted preprocessing
objects, trained models and all metric artefacts are serialised; the analyses
and figures in this paper are regenerated from those artefacts by released
scripts.

\section{Predictive Performance}
\label{sec:results}

\begin{table*}[t]
\caption{Test-set performance. All values recomputed from the serialised
experiment artefacts. AUC intervals are Hanley--McNeil; $\tau^{\star}$ is the
$F_1$-optimal threshold selected on training data. The generalisation gap
column is train minus test ROC-AUC.}
\label{tab:main}
\centering
\footnotesize
\begin{tabular}{@{}lcccccccc@{}}
\toprule
\textbf{Model} & \textbf{ROC-AUC} & \textbf{95\% CI} & \textbf{PR-AUC} &
$\boldsymbol{F_1}$ & \textbf{Precision} & \textbf{Recall} &
\textbf{Specificity} & \textbf{Gap}\\
\midrule
\textbf{Multi-scale ensemble} & \textbf{0.9539} & $[0.9462,\,0.9616]$ &
\textbf{0.9137} & \textbf{0.8397} & 0.9636 & \textbf{0.7440} & 0.9921 & $-0.0056$\\
\midrule
Random forest        & 0.9396 & $[0.9309,\,0.9484]$ & 0.8948 & 0.8356 & 0.9599 & 0.7398 & 0.9914 & $-0.0082$\\
XGBoost              & 0.9358 & $[0.9268,\,0.9448]$ & 0.8867 & 0.8241 & \textbf{0.9787} & 0.7117 & \textbf{0.9957} & $-0.0079$\\
Residual network     & 0.9306 & $[0.9213,\,0.9399]$ & 0.8764 & 0.8094 & 0.9273 & 0.7180 & 0.9843 & $+0.0063$\\
Deep \& Cross network & 0.9251 & $[0.9155,\,0.9348]$ & 0.8653 & 0.7933 & 0.9076 & 0.7046 & 0.9800 & $-0.0027$\\
TabNet (pure)        & 0.9218 & $[0.9120,\,0.9317]$ & 0.8608 & 0.7946 & 0.9180 & 0.7004 & 0.9825 & $+0.0051$\\
TabNet + tokenizer   & 0.9197 & $[0.9098,\,0.9297]$ & 0.8462 & 0.7747 & 0.8787 & 0.6927 & 0.9733 & $+0.0012$\\
Logistic regression  & 0.8663 & $[0.8538,\,0.8787]$ & 0.6956 & 0.6454 & 0.6248 & 0.6674 & 0.8881 & $-0.0027$\\
\bottomrule
\end{tabular}
\end{table*}

\begin{figure*}[t]
\centering
\includegraphics[width=\textwidth]{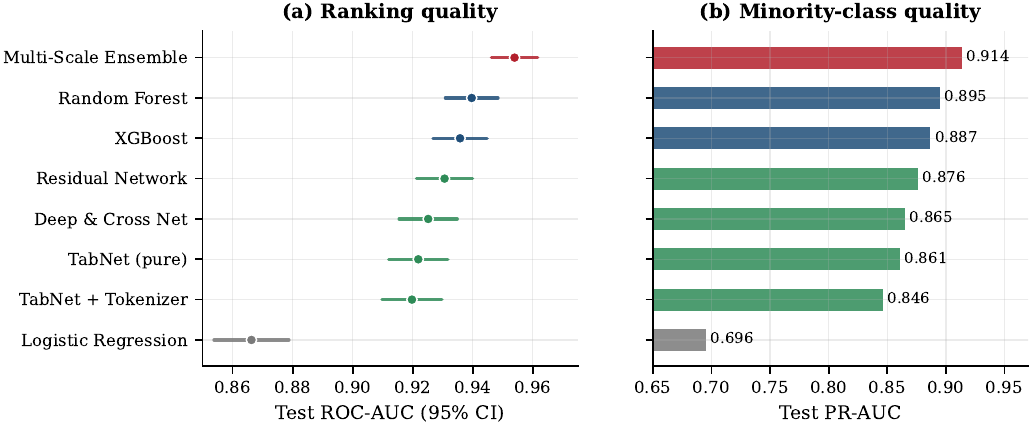}
\caption{Test performance with uncertainty. (a) ROC-AUC with Hanley--McNeil
$95\%$ intervals. The ensemble's interval is disjoint from those of the neural
architectures and overlaps the random forest's only marginally. (b) PR-AUC,
the more informative summary under a $3.58{:}1$ imbalance.}
\label{fig:perf}
\end{figure*}

Table~\ref{tab:main} and Fig.~\ref{fig:perf} report the full comparison.

\textbf{The ordering reproduces the tabular literature.} Random forest
($0.9396$) and XGBoost ($0.9358$) beat every neural architecture we trained.
The best network, the residual MLP, reaches $0.9306$; TabNet's learnable
tokenizer variant is the weakest of the deep models at $0.9197$, so the
additional feature-tokenisation machinery cost more than it returned on a
$17$-dimensional design matrix. This is consistent with the finding that
tree ensembles retain an advantage on typical tabular
data~\cite{grinsztajn2022, shwartzziv2022}. It is worth stating plainly
because it undercuts a common framing: the deep architectures here are not
where the performance comes from.

\textbf{The ensemble improves on the best single model, and the improvement is
statistically supported.} Against the random forest,
$\Delta\mathrm{AUC}=0.0143$. Under the conservative independence assumption of
Section~\ref{sec:protocol}, $z=2.40$ and $p=0.016$. Under more realistic
correlation the margin is far more decisive: at $\rho=0.5$, $p=7.3\times
10^{-4}$; at $\rho=0.8$, $p=1.3\times10^{-7}$. Against the best neural
architecture, $\Delta\mathrm{AUC}=0.0233$ with $p=1.6\times10^{-4}$ even under
the conservative assumption. The stacking gain is therefore real rather than
noise, which is the claim H1 makes and the evidence supports.

\textbf{Overfitting is controlled.} The train--test ROC-AUC gap does not exceed
$0.0082$ in absolute value for any model, and is negative for the ensemble
($-0.0056$, \ie{} test exceeds train). For a system with five base learners and
a meta-network this is the reassuring outcome, and it is attributable to the
out-of-fold protocol and early stopping rather than to luck.

\textbf{PR-AUC separates the models more sharply than ROC-AUC.} The ensemble's
$0.9137$ against the random forest's $0.8948$ is a $0.0189$ gap, larger than
the corresponding ROC gap. Under imbalance this is the more relevant
comparison, since PR-AUC is sensitive to performance on the minority class that
motivates the exercise.

\section{Operating Point and Cost}
\label{sec:cost}

\begin{table}[t]
\caption{Test confusion matrices and Wilson $95\%$ intervals at each model's
$F_1$-optimal threshold. $n=6{,}517$ ($1{,}422$ defaults).}
\label{tab:confusion}
\centering
\footnotesize
\setlength{\tabcolsep}{3.5pt}
\begin{tabular}{@{}lrrrrcc@{}}
\toprule
\textbf{Model} & \textbf{TN} & \textbf{FP} & \textbf{FN} & \textbf{TP} &
\textbf{Recall (95\% CI)} & \textbf{Prec. (95\% CI)}\\
\midrule
Ensemble        & 5055 & 40 & 364 & 1058 & $[0.721,\,0.766]$ & $[0.951,\,0.973]$\\
Random forest   & 5051 & 44 & 370 & 1052 & $[0.716,\,0.762]$ & $[0.947,\,0.970]$\\
XGBoost         & 5073 & 22 & 410 & 1012 & $[0.688,\,0.735]$ & $[0.968,\,0.986]$\\
Residual net    & 5015 & 80 & 401 & 1021 & $[0.694,\,0.741]$ & $[0.911,\,0.941]$\\
Logistic reg.   & 4525 & 570 & 473 & 949 & $[0.643,\,0.691]$ & $[0.600,\,0.649]$\\
\bottomrule
\end{tabular}
\end{table}

\begin{figure}[t]
\centering
\includegraphics[width=\columnwidth]{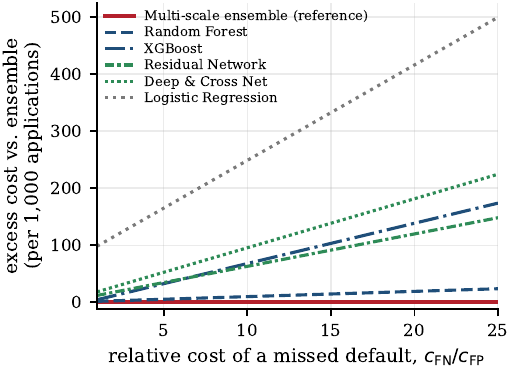}
\caption{Excess cost-weighted loss relative to the proposed ensemble at each
model's $F_1$-optimal threshold, as the relative cost of a missed default
varies. The ensemble is best throughout, but its margin over a tuned random
forest is small: under $2\%$ across the plotted range. The visible separation
is between the ensemble and the \emph{neural} models, not between the ensemble
and the tree baselines.}
\label{fig:cost}
\end{figure}

Table~\ref{tab:main} reports ranking quality. Deployment consumes decisions,
and the translation is not automatic.

Consider the confusion matrices in Table~\ref{tab:confusion}. At its selected
threshold the ensemble produces $364$ false negatives and $40$ false positives.
The random forest produces $370$ and $44$. The ensemble's entire operational
advantage over the strongest baseline is \emph{six} additional defaults caught
and \emph{four} fewer good applicants wrongly declined, out of $6{,}517$
decisions. The Wilson intervals for recall overlap almost completely
($[0.721,0.766]$ versus $[0.716,0.762]$).

To make this precise, let $c_{\mathrm{FN}}$ and $c_{\mathrm{FP}}$ be the costs
of a missed default and a wrongly declined applicant, and define expected cost
per thousand applications
\begin{equation}
C(\kappa) \;=\; \frac{1000}{n}\bigl(\mathrm{FP} + \kappa\,\mathrm{FN}\bigr),
\qquad \kappa = c_{\mathrm{FN}}/c_{\mathrm{FP}} .
\end{equation}
Fig.~\ref{fig:cost} plots the excess of each model over the ensemble. At
$\kappa=5$ the ensemble costs $285.4$ against the random forest's $290.6$, a
$1.80\%$ reduction; at $\kappa=10$, $564.7$ against $574.5$ ($1.71\%$); at
$\kappa=20$, $1123.2$ against $1142.2$ ($1.67\%$). The reduction is remarkably
stable in $\kappa$ and remarkably small.

Two conclusions follow, and they pull in opposite directions.

First, the honest one: \emph{a $0.0143$ AUC gain does not buy $0.0143$ worth of
anything at a fixed threshold.} ROC-AUC integrates over all thresholds; a
deployed system occupies one. Papers that report only AUC improvements,
including this one as originally drafted, invite the reader to infer an
operational benefit that the confusion matrix does not support. We would rather
state the limitation than let the abstract carry an implication we cannot
defend.

Second, the qualification: all models here are evaluated at \emph{their own}
$F_1$-optimal thresholds, and $F_1$ weights precision and recall equally, which
corresponds to $\kappa \approx 1$. No lender believes that. At $\kappa=10$,
every model in Table~\ref{tab:confusion} is operating well away from its
cost-optimal point, and the ensemble's superior ranking gives it more room to
move: a better-ordered score list yields a better precision--recall frontier to
select from. Threshold selection under an explicit cost model is, on this
evidence, a higher-leverage intervention than further architecture search, a
point we return to in Section~\ref{sec:discussion}.

\section{What the Attribution Layer Actually Shows}
\label{sec:explanations}

\begin{figure}[t]
\centering
\includegraphics[width=\columnwidth]{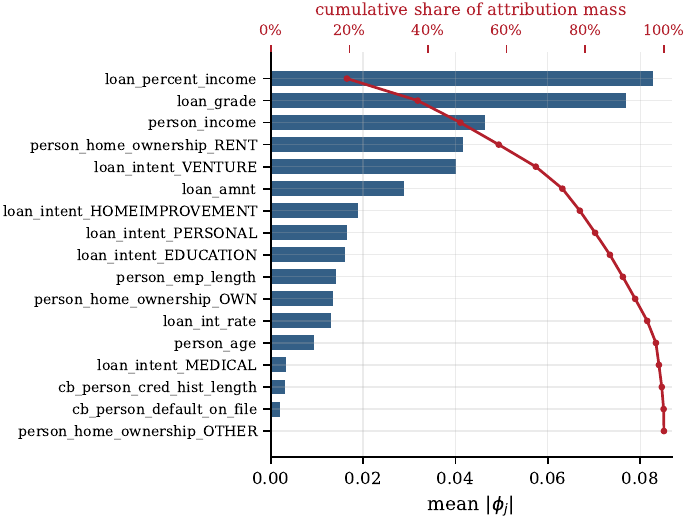}
\caption{Global attribution profile of the ensemble over $100$ explained test
instances (bars, left axis: mean $|\phi_j|$; line, top axis: cumulative share).
Attribution is diffuse: the top five features carry $67.4\%$ of the mass and
eight are needed to reach $80\%$. Note the position of
\texttt{loan\_int\_rate} (12th) and \texttt{cb\_person\_default\_on\_file}
(16th), both of which are leading predictors in the logistic baseline.}
\label{fig:shapglobal}
\end{figure}

\subsection{The ensemble's risk representation is not the linear model's}

Fig.~\ref{fig:shapglobal} gives the global profile. \fv{loan_percent_income}
leads at $\bar\phi = 0.0827$, followed by \fv{loan_grade} ($0.0768$) and
\fv{person_income} ($0.0464$).

The interesting result is what is \emph{absent}. In the L1-regularised logistic
baseline, the largest positive coefficients belong to
\fv{loan_percent_income}, \fv{loan_int_rate} and the bureau default flag. In
the ensemble, \fv{loan_int_rate} ranks twelfth ($0.0131$) and
\fv{cb_person_default_on_file} sixteenth of seventeen ($0.0020$).

This is not a contradiction, and it is not a defect. \fv{loan_grade} is an
internal rating that is itself a function of interest rate and bureau history:
grade determines price. A linear model, unable to represent that dependence,
must recover the signal from the constituent variables. The ensemble
represents the composite directly and the constituents become redundant. The
practical consequence is specific and easy to get wrong: a compliance narrative
generated from ensemble attributions will not mention prior defaults, even
though prior defaults demonstrably influence the outcome through the grade.
An adverse-action notice that omits them may be accurate about the model and
still misleading about the decision. Attribution to a feature is not attribution
to a cause, and collinear composites make the gap operationally visible.

\subsection{Attribution is diffuse, not concentrated}

The earlier draft asserted that the top five features account for
``almost $75\%$'' of attribution mass. Recomputed from the stored SHAP matrix,
the figure is $67.4\%$; eight of seventeen features are required to reach
$80\%$. Locally the picture is similar: the median instance requires six
features to account for $80\%$ of its own $|\phi|$ mass (IQR $5$--$6$).

This matters directly for the narrative layer, which is handed only the top
four drivers. Averaged over the explained instances, those four carry $72.3\%$
of an instance's attribution mass, so the prompt discards more than a quarter
of the evidence before the language model sees it. The truncation is a
defensible engineering choice, since shorter prompts are easier to constrain,
but it is a lossy summary and the residual is not negligible.

\subsection{Cross-model agreement is high; SHAP--LIME agreement is not}

\begin{figure}[t]
\centering
\includegraphics[width=\columnwidth]{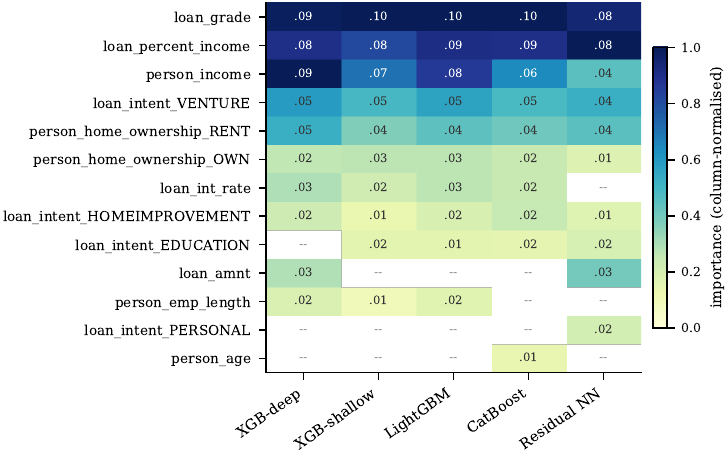}
\caption{Per-base-learner SHAP importances, normalised by column maximum;
``--'' marks a feature outside that learner's top ten. Mean pairwise
overlap@10 is $0.86$ and mean Kendall $\tau$ is $0.78$. \texttt{loan\_grade}
and \texttt{loan\_percent\_income} appear in every learner's top three.}
\label{fig:crossmodel}
\end{figure}

\begin{figure}[t]
\centering
\includegraphics[width=\columnwidth]{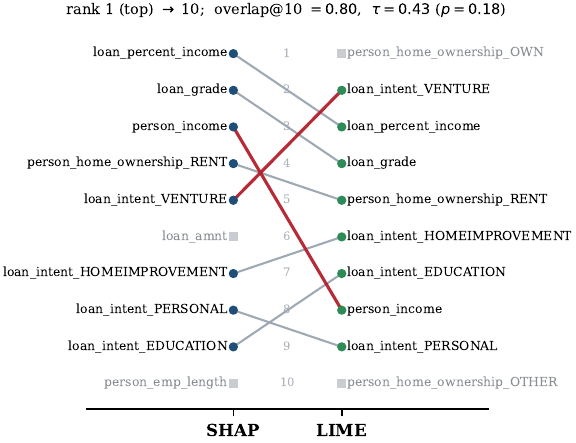}
\caption{Global rankings under SHAP and LIME. Grey squares mark features
present in only one ranking; red lines mark rank shifts of three or more.
The methods largely agree on set membership (overlap@10 $=0.80$) and disagree
on order ($\tau=0.43$, $p=0.18$), including at the very top: LIME's
first-ranked feature is SHAP's eleventh.}
\label{fig:ranks}
\end{figure}

Two agreement questions are often conflated. They have different answers here.

\textbf{Across base learners, agreement is genuinely high} (Fig.~\ref{fig:crossmodel}).
Mean pairwise overlap@10 is $0.860$ and mean Kendall $\tau$ is $0.776$;
\fv{loan_grade} and \fv{loan_percent_income} appear in all five learners' top
three, and seven features appear in all five top-ten lists. Disagreement is
concentrated in low-frequency one-hot levels: \fv{loan_intent_VENTURE} has a
coefficient of variation of $0.833$ across learners, \fv{cb_person_default_on_file}
$0.695$. So the claim that the ensemble's headline drivers are architecture-independent
is supported.

\textbf{Across attribution methods, agreement is weaker than the framework
assumes} (Fig.~\ref{fig:ranks}). SHAP and LIME share $8$ of $10$ top features
(overlap@10 $=0.80$), but rank correlation on the common set is only
$\tau=0.429$ ($p=0.179$), Spearman $\rho=0.524$ ($p=0.183$), neither
significant at $n=8$. Overlap@3 is $1/3$. LIME's top-ranked feature,
\fv{person_home_ownership_OWN}, is eleventh under SHAP; SHAP's top feature is
LIME's third.

Hypothesis H4 as originally stated, namely that SHAP and LIME show high
agreement on top features and that their combination is therefore more robust,
is supported for \emph{membership} and unsupported for \emph{ordering}. This is a
material qualification, because the fusion step averages normalised scores and
then truncates to the top four. Averaging two rankings that disagree on order
produces a third ranking that is not the consensus of anything; it is an
artefact of the normalisation. Where the methods agree on the set but not the
order, a set-level fusion (intersection or union with equal weight) would be
more defensible than a score-level average.

\subsection{Attribution sign is unstable for several leading features}

\begin{figure*}[t]
\centering
\includegraphics[width=\textwidth]{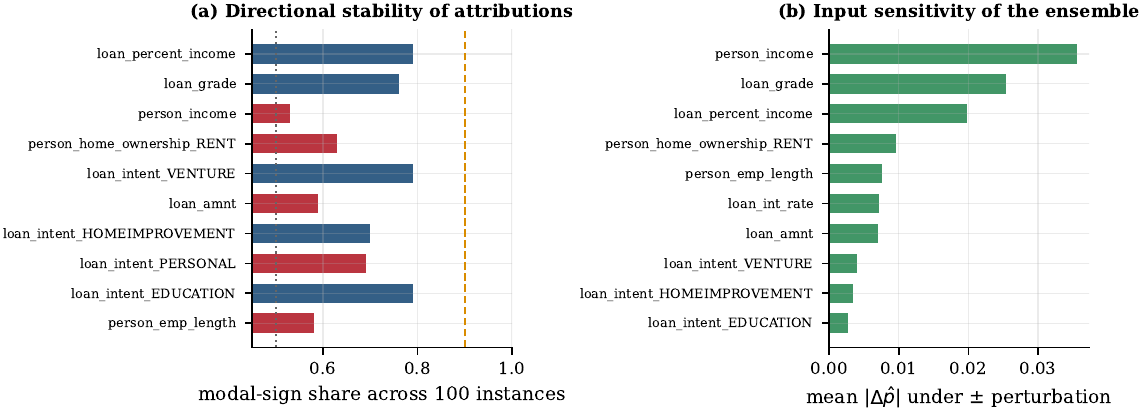}
\caption{(a) Directional stability: share of the $100$ explained instances on
which a feature's attribution takes its modal sign. Bars below $0.7$ (red)
indicate features whose risk direction routinely flips across applicants; a
value near $0.5$ means the direction is near-arbitrary at the population level.
(b) Sensitivity of the ensemble's output to bounded perturbation of each input.
The two panels disagree about \texttt{person\_income}, which is the most
sensitive input yet the least directionally stable.}
\label{fig:stability}
\end{figure*}

A narrative asserts direction: it says a factor raised or lowered risk. That
proposition is only meaningful if the attribution's sign is stable enough to
support it. Fig.~\ref{fig:stability}(a) measures this as the share of explained
instances on which each feature takes its modal sign.

The results are uneven. \fv{loan_percent_income} and
\fv{loan_intent_VENTURE} take a consistent sign on $79\%$ of instances and
\fv{loan_grade} on $76\%$, high but not decisive. \fv{person_income} sits at
$0.53$, statistically indistinguishable from a coin flip; \fv{loan_amnt} at
$0.59$; \fv{person_home_ownership_RENT} at $0.63$.

Some of this is legitimate interaction: income raises risk when paired with a
large principal and lowers it otherwise, and a model that captures that is
doing its job. But it has a consequence the framework must respect. Any
statement of the form ``the applicant's income increased their risk'' is a
claim about one instance that cannot be generalised, and the fusion step, which
averages \emph{normalised magnitudes} and discards sign, provides the language
model with no signal about how reliable the direction is.
Fig.~\ref{fig:stability}(b) sharpens the point: \fv{person_income} is the input
to which the ensemble's output is \emph{most} sensitive
($\overline{|\Delta \hat p|} = 0.0356$) and simultaneously the one whose
attribution direction is least stable.

This is the mechanism we see fail in the next section.

\section{Fidelity Audit of the Narrative Layer}
\label{sec:fidelity}

\subsection{Why an audit is needed}

The pipeline's grounding argument is entirely structural: the prompt contains
only computed drivers, the model is told to translate rather than infer, and
decoding is deterministic. Determinism is not fidelity. A greedy decode
produces the same output every time; it does not produce a \emph{correct} one.

We therefore define a minimal post-generation audit. Given an instance with
fused drivers $\{(j, \bar a_j)\}$ and generated narrative $y$:

\begin{itemize}
\item \textbf{Membership.} Every feature named in $y$ must appear in the
driver set supplied in the prompt.
\item \textbf{Direction.} For every named feature, the direction asserted in
$y$ (raises / lowers risk) must match $\operatorname{sign}(\bar a_j)$.
\item \textbf{Coverage.} The prompt's top driver by $|\bar a_j|$ should be
named.
\end{itemize}

These are necessary conditions, not sufficient ones, and they are cheap: all
three are checkable automatically against data already in memory.

\subsection{An audited failure}

Table~\ref{tab:fidelity} reports the audit on the worked case distributed with
the reference implementation: an applicant aged $34$, income \$92{,}000,
renting, requesting \$92{,}000 for a stated venture purpose at grade B and
$13.5\%$, with no bureau default. The ensemble returns $\hat p = 0.9402$,
band \textsc{high}.

The generated narrative reads:

\begin{quote}\itshape\small
``The risk level for this borrower is high due to their age, income, and home
ownership status. Their intent to take out a venture loan also increases the
risk. This case is similar to other high-risk cases in terms of income and home
ownership status. However, the intent to take out a venture loan is a unique
factor that sets this case apart. Based on these factors, it is recommended to
closely monitor this borrower's credit history and consider additional risk
mitigation measures.''
\end{quote}

It is fluent, appropriately hedged, correctly identifies the risk band, and
would pass casual review. It is also wrong in a specific and consequential way.

\begin{table}[t]
\caption{Fidelity audit of the worked case. Attributions are the fused
SHAP/LIME impacts supplied in the prompt; the assertion column is what the
generated narrative claims. Three of four checkable propositions invert the
sign of the evidence.}
\label{tab:fidelity}
\centering
\footnotesize
\setlength{\tabcolsep}{4pt}
\begin{tabular}{@{}lrll@{}}
\toprule
\textbf{Feature} & \textbf{Impact} & \textbf{Evidence} & \textbf{Narrative asserts}\\
\midrule
\texttt{loan\_amnt}          & $+0.3297$ & raises risk & \emph{not mentioned}\\
\texttt{...\_ownership\_RENT} & $+0.1179$ & raises risk & raises risk \checkmark\\
\texttt{...\_ownership\_OWN}  & $+0.0815$ & raises risk & \emph{not mentioned}\\
\midrule
\texttt{loan\_intent\_VENTURE} & $-0.0720$ & \textbf{lowers} risk & raises risk \ding{55}\\
\texttt{person\_income}        & $-0.0569$ & \textbf{lowers} risk & raises risk \ding{55}\\
\texttt{person\_age}           & $-0.0092$ & \textbf{lowers} risk & raises risk \ding{55}\\
\bottomrule
\end{tabular}
\end{table}

Three failures are visible in Table~\ref{tab:fidelity}:

\begin{enumerate}
\item \textbf{Sign inversion.} The narrative asserts that age, income and
venture intent \emph{increase} risk. The attributions score all three as
risk-\emph{reducing} ($-0.0092$, $-0.0569$, $-0.0720$). The model did not
misjudge magnitude; it reversed direction, and did so for the majority of the
features it chose to name.

\item \textbf{Coverage failure.} The dominant driver, \fv{loan_amnt} at $+0.3297$,
is never mentioned. It carries roughly three times the next largest impact and
is the one factor that plainly explains a $94\%$ default probability on a loan
equal to annual income.

\item \textbf{Membership violation.} \fv{person_age} is asserted as a risk
factor although it was not among the four drivers supplied.
\end{enumerate}

The narrative additionally attributes salience to venture intent
(``a unique factor that sets this case apart'') that the evidence assigns to
principal size.

\subsection{Diagnosis}

Three properties of the surrounding system make this outcome likely rather
than accidental.

\textbf{The prompt discards sign reliability.} Fusion averages normalised
magnitudes. The language model receives a signed impact but nothing about how
stable that sign is, and Section~\ref{sec:explanations} shows that for
\fv{person_income} it is barely stable at all ($0.53$).

\textbf{Prior knowledge overrides supplied evidence.} A 2.7B model has strong
priors about credit: high income is good, venture lending is speculative,
young borrowers are risky. When the supplied attributions contradict those
priors, as they do here since \emph{this} applicant's income and stated purpose
reduce risk relative to the population, the prior wins. The
instruction to translate rather than infer does not bind.

\textbf{Nothing checks the output.} Every control is upstream. The system has
no mechanism that could have caught a sign inversion, which is why a
deterministic decode produced a reproducibly wrong narrative rather than an
occasionally wrong one.

\subsection{Scope of this finding}

We are explicit about what this is and is not. It is a documented,
reproducible failure of a specific model under a specific prompt on a specific
instance, verified against the attributions supplied in the same run. It is
\emph{not} an estimate of failure frequency: producing one would require
generating and adjudicating narratives across a labelled sample, which is
future work we describe in Section~\ref{sec:limitations}. A single audited
failure cannot establish a rate. It is, however, sufficient to refute the
claim the earlier draft made, that the narratives ``accurately reflect the
underlying feature contributions'', and sufficient to establish that input-side
constraints alone do not guarantee grounding.

Our recommendation follows directly from the audit definition: the three checks
above are cheap, automatic, and would have blocked this output. A narrative
layer in a regulated setting should run them and fall back to a deterministic
template when they fail. This is the two-stage generate-then-verify pattern
argued for in~\cite{twostage2026}, and our result is a concrete instance of why
the second stage is not optional.

\section{Trustworthiness Diagnostics}
\label{sec:trust}

\begin{table}[t]
\caption{Trustworthiness diagnostics against the acceptance thresholds
specified by our own evaluation protocol. Three of four fail.}
\label{tab:trust}
\centering
\footnotesize
\setlength{\tabcolsep}{4pt}
\begin{tabular}{@{}llcc@{}}
\toprule
\textbf{Diagnostic} & \textbf{Value} & \textbf{Threshold} & \textbf{Verdict}\\
\midrule
Calibration (ECS, weighted)   & $0.1171$ & $<0.05$ & \textcolor{hl}{fail}\\
\quad unweighted              & $0.2262$ & n/a     & n/a\\
Perturbation stability (DPD)  & $0.0780$ & $<0.05$ & \textcolor{hl}{fail}\\
Local explanation consistency & $\overline{\mathrm{CV}}=1.1495$ & n/a & \textcolor{hl}{low}\\
Group disparity (EOD)         & $0.0640$ & $<0.10$ & pass\\
\bottomrule
\end{tabular}
\end{table}

\begin{figure}[t]
\centering
\includegraphics[width=\columnwidth]{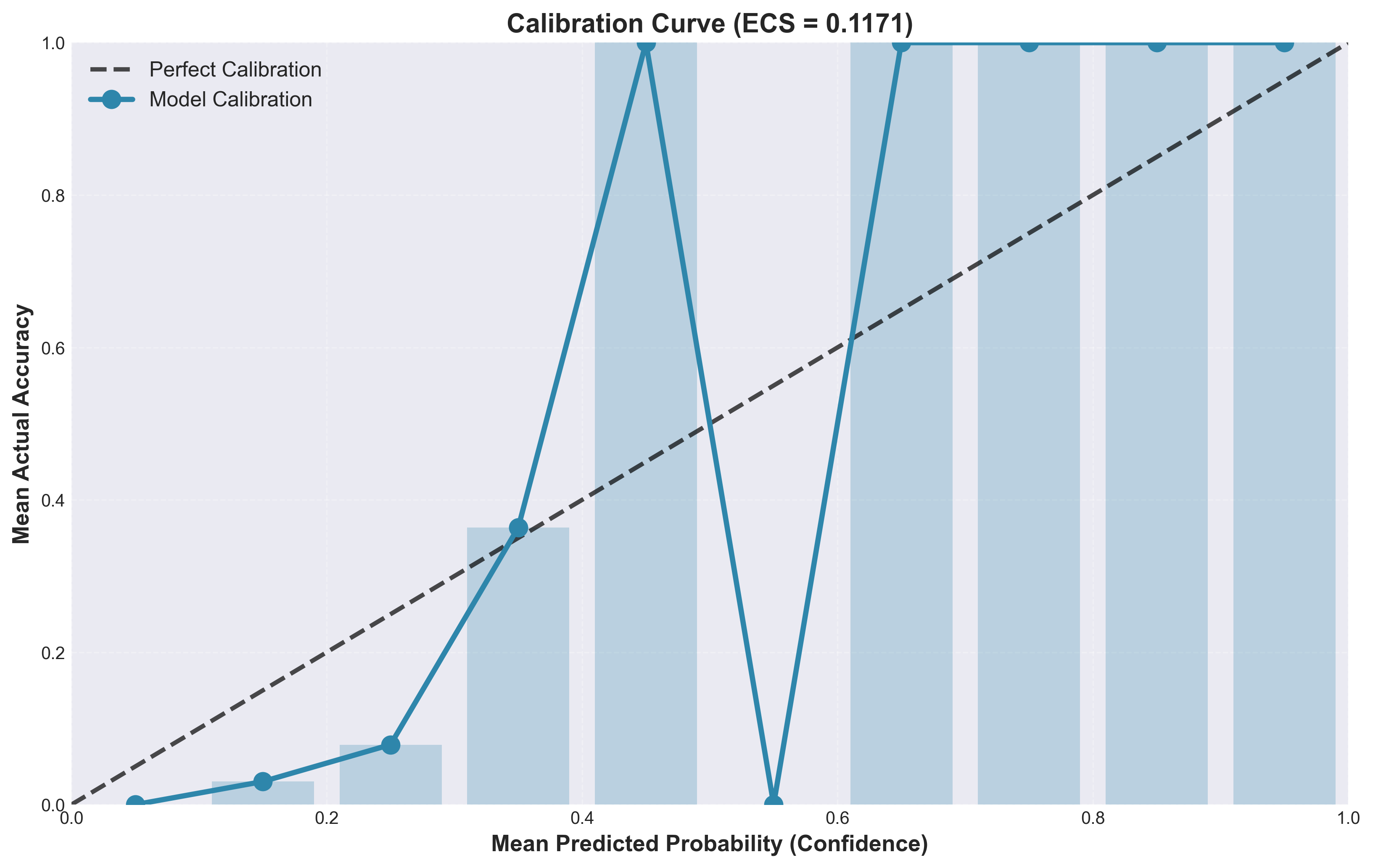}
\caption{Reliability diagram for the ensemble ($n=500$, $10$ bins,
$\mathrm{ECS}=0.1171$). The model is over-confident below $\hat p \approx 0.3$
and erratic above it. The upper bins are sparsely populated, so the extreme
excursions should be read as high-variance rather than as evidence of
systematic bias. The low-probability region, where most applications fall, is
well populated and consistently miscalibrated.}
\label{fig:calib}
\end{figure}

Ranking quality is one property of a scoring model. A credit system uses the
probability itself: to price risk, to set the \textsc{high}/\textsc{moderate}/
\textsc{low} band that appears in the narrative prompt, and to aggregate
expected loss. Table~\ref{tab:trust} evaluates the ensemble against the acceptance
thresholds our own protocol declares.

\textbf{Calibration fails.} The expected calibration score is $0.1171$
weighted and $0.2262$ unweighted, against a $0.05$ acceptance threshold.
Fig.~\ref{fig:calib} shows systematic over-confidence in the low-probability
region where the overwhelming majority of applications sit. This is the
expected consequence of two design choices: focal loss deliberately distorts
the probability scale in exchange for better minority-class ranking, and
modern neural networks are miscalibrated by default~\cite{guo2017calibration}.
The remedy is standard and cheap, namely Platt scaling or isotonic regression
on a held-out fold~\cite{niculescu2005}, and we did not apply it. We flag this
rather than omit it because miscalibrated probabilities propagate directly into
the risk band shown to the loan officer.

\textbf{Perturbation stability fails.} Under bounded input perturbation
(strength $0.01$, $10$ repetitions) the mean absolute prediction shift is
$0.0780$ ($\sigma=0.0595$, max $0.3284$), against a $0.05$ threshold. A $1\%$
input perturbation moving the output by nearly $8$ points is a material
sensitivity for a system whose inputs are self-reported.

\textbf{Local explanation consistency is low.} Clustering the explained
instances into five groups and measuring within-cluster dispersion of
attribution vectors gives a mean coefficient of variation of $1.1495$, so
dispersion exceeds the mean. Similar applicants receive materially different
explanations. This is consistent with the sign instability of
Section~\ref{sec:explanations} and with LIME's documented sampling
variance~\cite{alvarezmelis2018}.

\textbf{Group disparity passes, with an important caveat.} Equalised odds
difference~\cite{hardt2016} is $0.0640$ overall, below the $0.10$ threshold,
with the largest gap on the low-debt-ratio split ($0.0839$). The caveat is that
the groups examined (grade presence, high income, low debt ratio) are
\emph{model-derived proxies}, not protected attributes. The dataset contains no
race, gender or ethnicity fields, so no statement about discrimination on
protected characteristics can be made from these numbers, and we make none.
Reporting a passing EOD without that qualification would be the most misleading
thing in this paper.

\section{Discussion}
\label{sec:discussion}

\subsection{Where the gains are, and are not}

Read together, our results describe a system whose two halves succeed
differently.

The predictive half works, modestly. Multi-scale stacking beats the best single
model by a margin that survives a deliberately unfavourable significance test,
generalises without overfitting despite six trainable components, and improves
PR-AUC by more than it improves ROC-AUC, which is the right direction under
imbalance. But its operational advantage at the deployed threshold is six
defaults in six thousand decisions. An institution weighing five base learners,
a meta-network and the attendant governance burden against a single random
forest should know that number before committing.

The explanatory half is where the interesting failure is, and it is not where
the literature usually looks. The attribution layer is sound in construction:
explaining the composed pipeline rather than a component is the right choice,
and cross-learner agreement confirms the headline drivers are not artefacts.
What breaks is the last step, the one added specifically to make the system
usable by non-specialists. That step inherits every instability upstream of it
(diffuse attribution, unstable signs, method disagreement on ordering),
compresses them into four numbers, and hands them to a model with strong
independent priors and no obligation to defer.

\subsection{Implications for LLM-based explanation interfaces}

Three points generalise beyond this system.

\textbf{Constrained prompting is a necessary but not sufficient control.} Our
prompt does everything the literature recommends. It still produced sign
inversions. The reason is structural: prompt constraints shape the input
distribution and cannot bind the output. Any deployment that treats prompt
design as the safety mechanism is relying on a control that has no
enforcement.

\textbf{Verification is cheap and should be mandatory.} Membership, direction
and coverage checks require no model calls, no labels and no human review. They
would have blocked our failure case. In a regulated setting, an unverified
narrative is a liability whose cost is bounded by the size of the credit
decision, and the check costs microseconds.

\textbf{Explanation quality inherits attribution quality.} A narrative cannot
be more faithful than the attributions it summarises. Reporting sign stability
and inter-method rank agreement alongside importance rankings should be
standard, because these properties, not mean $|\phi|$, determine whether a
directional claim about an individual applicant is supportable.

\subsection{Practical guidance}

For practitioners building similar systems, our results suggest an ordering of
effort that differs from the one we followed. Calibrate before deploying, since
miscalibrated probabilities corrupt both pricing and the risk bands that reach
human reviewers, and the fix is a single held-out fold. Optimise the threshold
against an explicit cost ratio rather than $F_1$; Fig.~\ref{fig:cost} indicates
this matters more than the choice among strong models. Verify narratives after
generation. And treat additional base learners as the last resort rather than
the first, since our fifth learner bought less than our threshold choice did.

\section{Limitations and Future Work}
\label{sec:limitations}

\textbf{Single dataset.} All results come from one public dataset with a fixed
$21.8\%$ default rate, no temporal structure and no macroeconomic covariates.
Whether the meta-learner's weighting generalises across imbalance regimes,
feature distributions or credit cycles is untested.

\textbf{The fidelity audit is a case study.} We document one reproducible
failure in detail. We do not estimate a failure rate. The natural extension is
to generate narratives for a stratified sample of several hundred test
instances and score them automatically against the three audit criteria, which
would yield per-criterion violation rates and support comparison across model
scales and prompt designs. We regard this as the single highest-value follow-up
and note that our audit definition makes it inexpensive.

\textbf{Correlated AUC comparison.} Paired score vectors were not retained, so
DeLong's test~\cite{delong1988} could not be run. Our conservative substitute
understates significance; re-running with retained scores would tighten the
comparison in our favour, which is the direction that requires no defence.

\textbf{No calibration stage.} We report miscalibration but do not correct it.
Adding Platt or isotonic scaling on a dedicated fold, then re-deriving the risk
bands, is straightforward and should precede any deployment.

\textbf{Fixed base hyperparameters.} Base configurations were set from prior
experiments and held fixed while the meta-learner was tuned. Joint optimisation
might improve the ensemble, at substantially higher cost.

\textbf{Static evaluation.} Batch prediction on a fixed split does not test
concept drift, which is the dominant failure mode of deployed credit models
across economic cycles.

\textbf{Protected attributes absent.} The dataset contains no protected
characteristics, so our disparity analysis uses proxies and cannot speak to
discrimination.

Beyond these, two directions follow naturally from our findings: fine-tuning or
constrained decoding that forces the narrative model to respect supplied signs,
and set-level rather than score-level attribution fusion, which
Section~\ref{sec:explanations} suggests would be better founded given that the
two methods agree on membership but not order.

\section{Conclusion}

We built a credit-risk system of the kind currently advocated in the
explainable-AI literature: a high-performing ensemble, model-agnostic
attribution over the composed pipeline, and a constrained language model to
render the attributions as prose. We then measured both halves against their
own claims.

The ensemble performs as intended. Multi-scale stacking of four
differently-regularised boosting learners and a residual network reaches test
ROC-AUC $0.9539$ and PR-AUC $0.9137$, improving on the best single model by a
margin that remains significant ($p=0.016$) under a test chosen to disadvantage
it, with no evidence of overfitting. We also show that this margin is worth
less operationally than it appears: under $2\%$ of cost-weighted loss at the
deployed threshold.

The narrative layer does not perform as claimed. Constrained prompting,
deterministic decoding and top-$k$ truncation did not prevent a generated
rationale from inverting the risk direction of three of the four factors it
named, omitting the dominant driver, and introducing a feature it was never
given. We trace this to measurable properties of the attribution layer beneath
it: attribution is diffuse ($67.4\%$ mass in five of seventeen features), SHAP
and LIME agree on membership but not order ($\tau=0.43$, $p=0.18$), and the
sign of the attribution for the ensemble's most sensitive input is near-random
across applicants ($0.53$). We report alongside these that the ensemble is
miscalibrated ($\mathrm{ECS}=0.117$) and sensitive to bounded input
perturbation ($\mathrm{DPD}=0.078$), both outside our own acceptance
thresholds.

The conclusion we draw is not that language models have no place in credit
explanation. It is that grounding must be verified rather than assumed, that the
verification is cheap, requiring only membership, direction and coverage checks
over data already in memory, and that a system which enforces it can make the
claim we could not: that the explanation given to the applicant matches the
evidence the model actually used.

\appendices
\section{Narrative Prompt Template}
\label{app:prompt}

The template below is reproduced verbatim from the deployed service. Bracketed
italics denote substituted values.

\vspace{2pt}
\noindent\fbox{\parbox{0.95\columnwidth}{\footnotesize\ttfamily\raggedright
You are an AI credit risk officer. Provide a clear, professional explanation.\\[2pt]
Prediction: \textit{[p]}\% default probability $\rightarrow$ Risk Level: \textit{[band]}\\[2pt]
Borrower Profile:\\
\textit{[raw feature: value, one per line]}\\[2pt]
Local Feature Drivers (for this specific case):\\
\textit{[feature: increases/decreases risk (impact: $\pm$x.xxx), top 4]}\\[2pt]
Global Context: Top globally important features are \textit{[top 5]}.\\
Most sensitive features (from analysis): \textit{[top 3]}.\\[2pt]
Provide a 2-3 sentence explanation: (1) State the risk level and key
factors, (2) Explain how this case compares to typical patterns,
(3) Give a recommendation.
}}
\vspace{2pt}

Generation uses the following decoding parameters.

\vspace{2pt}
\begin{center}\footnotesize
\begin{tabular}{@{}>{\ttfamily}l l@{}}
\toprule
\normalfont\textbf{Parameter} & \textbf{Value}\\
\midrule
max\_new\_tokens     & 256\\
min\_new\_tokens     & 64\\
do\_sample           & \texttt{False} (greedy)\\
repetition\_penalty  & 1.05\\
\bottomrule
\end{tabular}
\end{center}
\vspace{2pt}

The driver list carries signed impacts, so the direction that the audit in
Section~\ref{sec:fidelity} finds inverted was explicitly present in the prompt.


\end{document}